%% file: main.tex
\documentclass[journal]{IEEEtran}
\IEEEoverridecommandlockouts
\usepackage{cite}
\usepackage{amsmath,amssymb,amsfonts,physics, siunitx}
\usepackage{algorithmic}
\usepackage{graphicx}
\usepackage{textcomp}
\usepackage{xcolor}
\usepackage{tikz, enumitem, hyperref}
\usetikzlibrary{positioning}
\def\BibTeX{{\rm B\kern-.05em{\sc i\kern-.025em b}\kern-.08em
    T\kern-.1667em\lower.7ex\hbox{E}\kern-.125emX}}
\begin{document}

\title{Change Point Detection in Time Series Data using Autoencoders with a Time-Invariant Representation
%{\footnotesize \textsuperscript{*}Note: Sub-titles are not captured in Xplore and should not be used}
\thanks{This research received funding from the Flemish Government (AI Research Program) and from the European Research Council (ERC) under the European Union’s Horizon 2020 research and innovation programme (grant agreement No 802895). 
All authors are affiliated to Leuven.AI - KU Leuven institute for AI, B-3000, Leuven, Belgium.}
}

\author{
%\IEEEauthorblockN{Tim De Ryck}
% \IEEEauthorblockA{\textit{Seminar for Applied Mathematics} \\
% \textit{Department of Mathematics}\\
% \textit{ETH Zürich}\\
% Zürich, Switzerland \\
% deryckt@ethz.ch}
%\and
\IEEEauthorblockN{Tim De Ryck\IEEEauthorrefmark{1}\IEEEauthorrefmark{2}, Maarten De Vos\IEEEauthorrefmark{1}\IEEEauthorrefmark{3}, Alexander Bertrand\IEEEauthorrefmark{1}}\\
\IEEEauthorblockA{\IEEEauthorrefmark{1} \textit{STADIUS Center for Dynamical Systems, Signal Processing and Data Analytics,} \\
\textit{Department of Electrical Engineering (ESAT), KU Leuven, Belgium} \\
\textit{\IEEEauthorrefmark{2} Seminar for Applied Mathematics, Department of Mathematics, ETH Zürich, Switzerland}\\
\textit{\IEEEauthorrefmark{3} Department of Development and Regeneration, KU Leuven Belgium}
%\\alexander.bertrand@esat.kuleuven.be, maarten.devos@esat.kuleuven.be
}

}

\maketitle

\begin{abstract}
Change point detection (CPD) aims to locate abrupt property changes in time series data. Recent CPD methods demonstrated the potential of using deep learning techniques, but often lack the ability to identify more subtle changes in the autocorrelation statistics of the signal and suffer from a high false alarm rate. To address these issues, we employ an autoencoder-based methodology with a novel loss function, through which the used autoencoders learn a partially time-invariant representation that is tailored for CPD. The result is a flexible method that allows the user to indicate whether change points should be sought in the time domain, frequency domain or both. Detectable change points include abrupt changes in the slope, mean, variance, autocorrelation function and frequency spectrum. We demonstrate that our proposed method is consistently highly competitive or superior to baseline methods on diverse simulated and real-life benchmark data sets. Finally, we mitigate the issue of false detection alarms through the use of a postprocessing procedure that combines a matched filter and a newly proposed change point score. We show that this combination drastically improves the performance of our method as well as all baseline methods. 
\end{abstract}


\input{introduction}
\input{method}
\input{experiments}

\input{conclusion}

\bibliographystyle{IEEEtran}
\bibliography{references.bib, references2.bib}

\end{document}

%% file: introduction.tex
\section{Introduction}
In the era of big data, where Internet of Things (IoT) devices and other sensors provide endless data streams, the importance of time series analysis techniques can hardly be overestimated. One particular task, that has drawn attention from statistics and data mining communities for decades\cite{wald2004sequential,brodsky2013nonparametric, gustafsson2000adaptive, basseville1993detection}, is \textit{change point detection} (CPD): the detection of abrupt changes in the temporal evolution of time series data. Change point detection can be a goal in itself or  it can be used as a preprocessing tool to segment a time series in homogeneous segments (which can then be further analysed, clustered or classified). 
%In its latter form, CPD is closely related to anomaly detection, where normal segments alternate with usually short anomalous segments or outliers. 
Real-life applications of CPD include, but are not limited to, the analysis of climate data \cite{reeves_chen_wang_lund_lu_2007}, financial market data \cite{pepelyshev_polunchenko_2017, hsu_1982}, genetic data \cite{wang_wu_ji_wang_liang_2011} sensor network data \cite{tartakovsky2003quickest, tartakovsky2008asymptotically} and medical data \cite{michael1979automatic, malladi2013online}. 

CPD methods can be categorized according to many different criteria. It is common to make the distinction between online CPD, which provides real-time detections, and retrospective (offline) CPD, which provides more robust detections at the cost of needing more future data. In this paper, we focus on the second category. Many CPD algorithms compare past and future time series intervals by means of a dissimilarity measure. An alarm is issued when the two intervals are sufficiently dissimilar. A first group of methods defines this dissimilarity measure based on the difference in distribution of the two intervals. CUSUM and related methods \cite{basseville1993detection,shao2010testing} track changes in the parameter of a chosen distribution, the generalized likelihood ratio (GLR) procedure \cite{brandt, appel_brandt_1983} monitors the likelihood that both intervals are generated from the same distribution, and subspace methods \cite{ide2007change, kawahara2007change} measure the distance between subspaces spanned by the columns of an observability matrix. All these methods however strongly rely on the assumption that the time series data is generated using some parametric probability distribution (CUSUM), autoregressive model (GLR) or state-space model (subspace method). Bayesian online CPD \cite{Adams2007BayesianDetection} is another notable algorithm that depends on distributional assumptions. Unsurprisingly, the performance of these parametric methods heavily depends on how well the actual data follows the assumed model. Parameter-free alternatives are kernel density estimation \cite{csorgHo198820, brodsky2013nonparametric, Chang2019KernelModels} and the related density ratio estimation \cite{Kawahara2012SequentialEstimation, Liu2013Change-pointEstimation}. A more complete overview of CPD methods can be found in e.g. \cite{Truong2020SelectiveMethods, namoano2019online, Burg2020AnAlgorithms, Aminikhanghahi2017ADetection}. 

Following the successful application of deep learning techniques in anomaly detection, a promising approach for CPD was to base the dissimilarity measure on the distance between features automatically learned by an autoencoder \cite{Lee2018TimeLearning}. Main advantages of this approach are the absence of distributional assumptions and the ability of autoencoders to extract complex features from data in a cost-efficient way. There are however also some severe drawbacks. First, there are no guarantees that the distance between consecutive features reflects the actual dissimilarity of the intervals, i.e. features may vary significantly even in the absence of a change point. Second, the correlated nature of time series samples is not adequately leveraged by vanilla autoencoders, which makes it challenging to detect abrupt changes in the frequency domain. This is not uncommon in CPD literature \cite{basseville1993detection, Adams2007BayesianDetection, Chang2019KernelModels, cheng2020optimal}. Some methods explicitly focus on abrupt changes in the spectrum \cite{moskvina2003change, Chen2019AutomatedAnalysis}, thereby often ignoring changes in the time domain. Finally, the absence of a postprocessing procedure preceding the detection of peaks in the dissimilarity measure often leads to a high number of false positive detection alarms \cite{Cheng2020OnDetection}. 

Building on \cite{Lee2018TimeLearning}, we propose a new autoencoder-based CPD method using a partially time-invariant representation (TIRE) that aims to overcome the aforementioned concerns. 
%The rest paper is structured as follows. In Section \ref{sec:problem}, we clarify our CPD setting. In Section \ref{sec:experiments}, we describe all details of the proposed methods. 
Our main contributions can be summarized as follows. 
\begin{itemize}
    \item We propose a new CPD framework based on a novel adaptation of the autoencoder with a loss function that promotes time-invariant features. Through our choice of loss function, we aim for the autoencoder to learn a representation that is better suited for CPD. Based on this encoding, we define a dissimilarity measure to detect change points. We evaluate the performance of our algorithm on diverse simulated and real-life benchmark data sets and compare with other dissimilarity-measure-based CPD algorithms. 
    \item Whereas many change point algorithms assume the time series data to consist of independent identically distributed (iid) samples, we explicitly focus on non-iid data. We use the discrete Fourier transform to obtain temporally localized spectral information and propose an approach that combines time-domain and frequency-domain information. When domain knowledge is available, our approach allows the user to only focus on the time or frequency domain. 
    \item Finally, we propose a way of identifying change points from the dissimilarity measure data by applying the notion of topographic prominence \cite{llobera_2001} to the CPD setting. We emphasize the general importance of postprocessing steps in CPD through numerous experiments. 
\end{itemize}

%% file: method.tex
\section{Problem formulation}\label{sec:problem}
Let $\mathbf{X}$ be a $d$-channel time series of length $T$ for which there exist time stamps $0=T_0<T_1<\ldots <T_p=T$ such that every subsequence of the form $(\mathbf{X}[T_k+1], \ldots, \mathbf{X}[T_{k+1}])$ is a realisation of a discrete time weak-sense stationary stochastic (WSS) process, whereas the union of two such consecutive subsequences is not. 
% A discrete time stochastic process $\{\mathcal{X}_t\}_t$ with autocovariance function $\gamma(t_1,t_2) = \mathbb{E}[(\mathcal{X}_{t_1}--\mathbb{E}[\mathcal{X}_{t_1}])(\mathcal{X}_{t_2}-\mathbb{E}[\mathcal{X}_{t_2}])]$ is said to be \textit{weak-sense stationary} (WSS) if \cite{rangayyan_2015}
% \begin{enumerate}
%     \item $\mathbb{E}[\mathcal{X}_{t_1}]=\mathbb{E}[\mathcal{X}_{t_1+t_2}]$ for all $t_1,t_2$,
%     \item $\gamma(t_1,t_2) = \gamma(t_1-t_2,0)$ for all $t_1,t_2$,
%     \item $\mathbb{E}\left[\norm{\mathcal{X}_t}^2\right]<\infty$ for all $t$.
% \end{enumerate}
The time stamps $T_1, T_2, \ldots$ are referred to as \textit{change points}. The goal of \textit{change point detection} (CPD) is to estimate these change points without any prior knowledge on the number and the locations of the change points \cite{Cheng2020OnDetection}. The piecewise WSS assumption is not a strict prerequisite for the algorithm to work, but it does accurately summarize the kind of change points our proposed algorithm will be able to detect. Examples of violations of the WSS conditions, and therefore examples of change points we wish to detect, are changes in mean, variance and autocorrelation. Note that changes in the latter are also reflected in the frequency spectrum through the Wiener-Khinchin theorem \cite{wiener_1930, khintchine_1934}. 

We focus on CPD algorithms that are based on a \textit{dissimilarity measure}. Such methods calculate for every time stamp $t$ the dissimilarity between the windows $(\mathbf{X}[t-N+1], \ldots, \mathbf{X}[t])$ and $(\mathbf{X}[t+1], \ldots, \mathbf{X}[t+N])$, where $N$ is a user-defined window size. Our first main goal is to develop a CPD-tailored feature embedding and a corresponding dissimilarity measure $\mathcal{D}_t$, which peaks when the WSS restriction is violated. The nominal approach for identifying change points would then be to determine all local maxima and label each local maximum of which the height exceeds a user-defined detection threshold $\tau$ as a change point \cite{M-stat, Lee2018TimeLearning}. However, given a window size $N$, the width of this peak will theoretically be $2N-1$ time stamps, making it likely that noise will cause multiple detection alarms for each ground-truth change point \cite{Cheng2020OnDetection}. Our second objective is to mitigate the impact of this issue. 

\section{Autoencoder-based change point detection}\label{sec:method}
\subsection{Preprocessing}
Let $\mathbf{X}$ be a $d$-channel time series of length $T$, where we denote the $i$-th channel by $\mathbf{X}^i$. We first divide each channel into windows of size $N$, 
\begin{equation}\label{eq:window-size}
    \mathbf{x}^i_t = \begin{bmatrix}\mathbf{X}^i[t-N+1],\ldots, \mathbf{X}^i[t]\end{bmatrix}^T \in \mathbb{R}^N.
\end{equation}
We then combine for every $t$ the corresponding windows of each channel into a single vector, 
\begin{equation}\label{eq:td-windows}
    \mathbf{y}_t = \begin{bmatrix}(\mathbf{x}^1_t)^T,\ldots, (\mathbf{x}^d_t)^T\end{bmatrix}^T \in\mathbb{R}^{Nd}.
\end{equation}
Furthermore, we use the discrete Fourier transform (DFT) on each window $\mathbf{x}^i_t$ to obtain temporally localized spectral information. The length of the transformed window is then cropped to a predefined length $M$. Finally, the modulus of the transformed window is computed. Bundling all these transformations as a single mapping $\mathcal{F}:\mathbb{R}^N\to\mathbb{R}^M$, we obtain the frequency-domain counterpart of $\mathbf{y}_t$:
\begin{equation}\label{eq:fd-windows}
    \mathbf{z}_t = \begin{bmatrix}\mathcal{F}(\mathbf{x}^1_t)^T,\ldots, \mathcal{F}(\mathbf{x}^d_t)^T\end{bmatrix}^T\in\mathbb{R}^{Md}.
\end{equation}
\subsection{Feature encoding}
Building on \cite{Lee2018TimeLearning}, we use autoencoders (AEs) to extract features for change point detection from the time-domain (TD) windows $\{\mathbf{y}_t\}_t$. We expand the approach in \cite{Lee2018TimeLearning} by also extracting features from the frequency-domain (FD) windows $\{\mathbf{z}_t\}_t$ and through the proposal of a new loss function that explicitly promotes time-invariance of the features in consecutive windows. The latter is a relevant property in order to perform CPD based on a dissimilarity metric. 

An autoencoder is a type of artificial neural network that aims to learn a low-dimensional encoding (i.e. features) from a higher-dimensional input by reconstructing the input from the encoding as accurately as possible. It is often used as a dimension reduction technique and can be seen as a non-linear generalization of PCA \cite{goodfellow2016deep}. In its simplest form, an autoencoder consists of one hidden layer. The encoder maps the input $\mathbf{y}_t\in\mathbb{R}
^{Nd}$ (resp. $\mathbf{z}_t$) to its encoded form $\mathbf{h}_t\in\mathbb{R}^{h}$ as 
\begin{equation}
    \mathbf{h}_t = \sigma(\mathbf{W}\mathbf{y}_t+\mathbf{b}),
\end{equation}
where $\mathbf{W}$ is the weight matrix, $\mathbf{b}$ is the bias vector and $\sigma$ is a non-linear activation function that is applied element-wise. The decoder then maps the encoded representation back to the original input space, 
\begin{equation}
    \Tilde{\mathbf{y}}_t = \sigma'(\mathbf{W'}\mathbf{h}_t+\mathbf{b'}). 
\end{equation}
We choose $\sigma=\sigma'$ to be the hyperbolic tangent function, with as a consequence that each channel of the time series should be rescaled to the interval $[-1, 1]$. We use individual instead of joint rescaling to ensure that all channels have a comparable magnitude. The goal of the AE is then to minimize the difference between the input and the output, i.e. minimize $\norm{\mathbf{y}_t-\Tilde{\mathbf{y}}_t}$, by optimizing the choice of $\mathbf{W}, \mathbf{W'}, \mathbf{b}, \mathbf{b'}$. In \cite{Lee2018TimeLearning}, the learned features $\mathbf{h}_t$ are then used for CPD by measuring the dissimilarity between consecutive feature vectors ($\mathbf{h}_t$ vs. $\mathbf{h}_{t-1}$). However, the learned features $\mathbf{h}_t$ will then unavoidably also contain information that is not relevant for CPD (e.g. phase shift or noise information), which may generate large dissimilarities even when there is no actual change point. 

We try to remedy this by introducing the notions of \textit{time-invariant} and \textit{instantaneous features}. The idea is that features learned from consecutive windows are only useful for CPD when they are approximately equal to each other in the absence of a change point (e.g. mean, amplitude and frequency should not change much within a WSS segment). We will refer to them as \textit{time-invariant} features as they are aimed to be invariant over time within a WSS segment. All other information that is needed for a good reconstruction, but that may differ for consecutive windows, is aimed to be encoded in \textit{instantaneous} features. This then gives
\begin{equation}
    \mathbf{h}_t =  \begin{bmatrix}(\mathbf{s}_t)^T, (\mathbf{u}_t)^T\end{bmatrix}^T,
\end{equation}
where $\mathbf{s}_t\in\mathbb{R}^{s}$ are the time-invariant features and $\mathbf{u}_t\in\mathbb{R}^{h-s}$ are the instantaneous features. To obtain both a good reconstruction and time-invariant features, we propose to minimize the loss function
\begin{equation}\label{eq:training-loss}
\sum_t\left( \norm{\mathbf{y}_{t}-\Tilde{\mathbf{y}}_{t}}_2 +\lambda \norm{\mathbf{s}_{t}-\mathbf{s}_{t-1}}_2\right),%    \sum_t\left( \norm{\mathbf{y}_{t}-\Tilde{\mathbf{y}}_{t}}_2 +\frac{\lambda}{K}\sum_{k=1}^{K} \norm{\mathbf{s}_{t+k}-\mathbf{s}_{t+k-1}}_2\right),
\end{equation}
where $\lambda>0$ control the amount of regularization of the time-invariant features. Here we make the implicit assumption that the number of terms in \eqref{eq:training-loss} that correspond to a window containing a change point is very small compared to $T$. 

\begin{figure}
    \centering
    \includegraphics[width=\columnwidth]{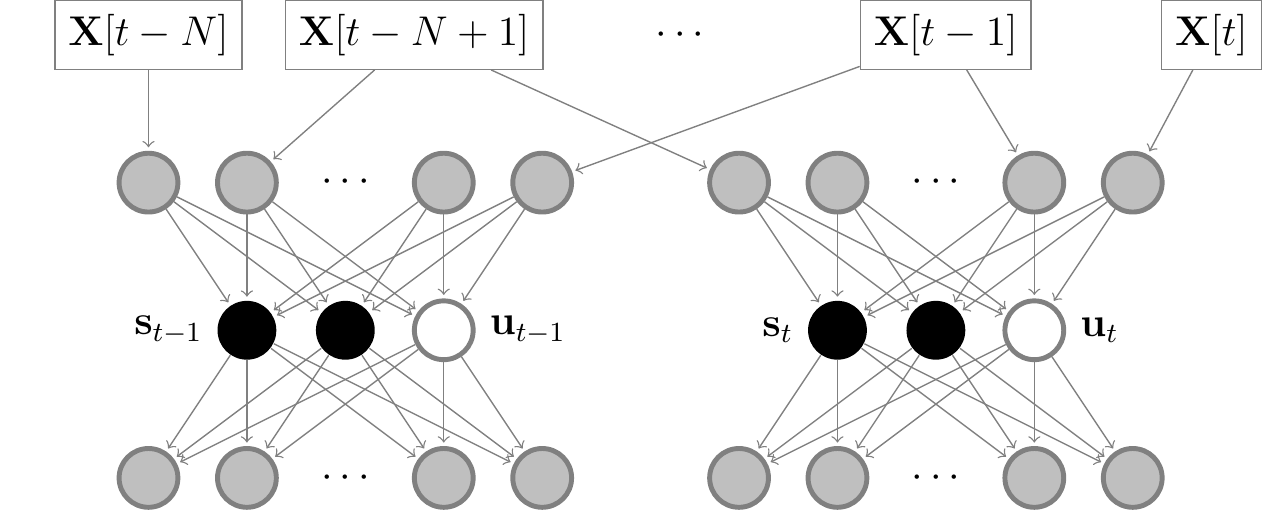}
    \caption{Visualization of time-invariant feature encoding for $K=1$. The TD autoencoder is shown two times, once with input $\mathbf{y}_{t-1}$ and once with input $\mathbf{y}_{t}$. The corresponding time-invariant features $\mathbf{s}_{t-1}$ and $\mathbf{s}_{t}$ are forced to be approximately equal because of the chosen loss function \eqref{eq:training-loss3}. Frequency-domain time-invariant features are obtained analogously.}
    \label{fig:pae}
\end{figure}

It is very uncommon in machine learning to directly minimize the loss function \eqref{eq:training-loss}, i.e. take all $t$ into account for every step of gradient descent. To improve convergence, it is advisable to first randomly partition all time stamps $t$ over $J$ smaller \textit{mini-batches} $\mathcal{T}_j$ \cite{masters2018revisiting}. The mini-batch stochastic gradient descent (SGD) version of minimizing \eqref{eq:training-loss} would then consist of updating the network parameters by calculating the gradient of 
\begin{equation}\label{eq:training-loss2}
\sum_{t\in\mathcal{T}_j}\left( \norm{\mathbf{y}_{t}-\Tilde{\mathbf{y}}_{t}}_2 +\lambda \norm{\mathbf{s}_{t}-\mathbf{s}_{t-1}}_2\right)
\end{equation}
for some $j$, followed by performing one gradient descent step and repeating this for all other mini-batches. Note that formulation \eqref{eq:training-loss2} would require to use time stamps from other batches, i.e. $t\in\mathcal{T}_j$ does not generally imply that $t-1\in\mathcal{T}_j$. However, we choose to generalize \eqref{eq:training-loss2}, and minimize the following loss function for each mini-batch, 
\begin{equation}\label{eq:training-loss3}
\sum_{t\in\mathcal{T}_j}\left( \norm{\mathbf{y}_{t}-\Tilde{\mathbf{y}}_{t}}_2 +\frac{\lambda}{K} \sum_{k=0}^{K-1}\norm{\mathbf{s}_{t-k}-\mathbf{s}_{t-k-1}}_2\right),
\end{equation}
where $K\in\mathbb{N}$. For $K=1$ this equation reduces to \eqref{eq:training-loss2}. For $K>1$, this approach has the advantage that now $K+1$ consecutive features are jointly and simultaneously considered during the computation of the gradient, resulting in an additional smoothing effect of the stochastic gradient in the direction of the minimization of the penalty term in \eqref{eq:training-loss}. Thereby further promoting the aimed time invariance of the features $\mathbf{s}_t$. It may help to think of \eqref{eq:training-loss3} as $K+1$ parallel autoencoders with identical weights and biases, where the $k$-th autoencoder receives $\mathbf{y}_{t+k-K-1}$ as input and where a subset of the latent variables (i.e. the time-invariant features) of the parallel autoencoders are forced to be close together to obtain a partially time-invariant representation (Figure \ref{fig:pae}). Note that even though the difference in formulation between \eqref{eq:training-loss} and \eqref{eq:training-loss3} impacts the training of the autoencoder, the resulting loss functions are essentially the same when summing over all $t$. 

To avoid that the autoencoder encodes all information in the unregularized instantaneous features, the number of instantaneous features should be taken as small as possible.  Depending on the data, one might add additional regularization terms to the loss function or use a more advanced type of autoencoder (e.g. weight regularized, deep/stacked, tied-weights, variational, recurrent autoencoder). In an entirely similar fashion, we train a second autoencoder on $\{\mathbf{z}_t\}_t$ with a similar loss function to obtain frequency-domain time-invariant features. We will use the superscripts TD and FD to distinguish between parameters and features corresponding to the time and frequency domain, respectively.  

\subsection{Postprocessing and peak detection}

In this section we first describe how to construct a dissimilarity measure that complies with the needs formulated in Section \ref{sec:problem} based on the time-invariant features from the previous section. Next, we discuss multiple methods to suppress the number of false positives when determining the detection alarms. 
\subsubsection{Postprocessing}
We first combine the TD and FD time-invariant features into a single time-invariant features vector, 
\begin{equation}\label{eq:combine_shared}
    \mathbf{s}_t = \begin{bmatrix}\alpha\cdot(\mathbf{s}^{\text{TD}}_t)^T,\beta\cdot(\mathbf{s}^{\text{FD}}_t)^T\end{bmatrix}^T,
\end{equation}
where $\alpha,\beta>0$ are parameters that control the relative contribution of the TD and FD time-invariant features. Next, we use a zero-delay weighted moving average filter to smoothen the time-invariant features, as small fluctuations in the features will affect the performance of the method. 
The moving average filtering operation can be described as follows, 
\begin{equation}\label{eq:smooth-features}
    \Tilde{\mathbf{s}}_t[i] = \sum_{k=-N+1}^{N-1} \mathbf{v}[N-k]\cdot \mathbf{s}_{t+k}[i], 
\end{equation}
% with 
% \begin{equation}
%     \mathbf{v}[k] =  \mathbf{v}[2N-k] \triangleq \frac{k}{N^2}\quad \text{for } 1\leq k\leq N. 
% \end{equation}
with $\mathbf{v}[k] = \mathbf{v}[2N-k] \triangleq k/N^2$ for $1\leq k\leq N$, where $N$ is the window size as defined in \eqref{eq:window-size}, resulting in a triangular shaped weighting window.
% First define the impulse response $\mathbf{v}\in\mathbb{R}^{2N-1}$ as
% \begin{equation}\label{eq:filter}
%     \mathbf{v}[k] =  \mathbf{v}[2N-k] = \frac{k}{N^2}\quad \text{for } 1\leq k\leq N. 
% \end{equation}
% We then apply the filter on each feature, 
We use edge value padding in order for the equation to be defined for all $t$. 
%A good dissimilarity measure $\mathcal{D}$ should be scale equivariant and invariant to a translation of the learned features. 
We then propose the following definition for the dissimilarity measure $\mathcal{D}$: 
\begin{equation}\label{eq:dissimilarity-measure}
    \mathcal{D}_t =
    %\mathcal{D}( \Tilde{\mathbf{s}}_t,\Tilde{\mathbf{s}}_{t+N}) =
    \norm{\Tilde{\mathbf{s}}_t- \Tilde{\mathbf{s}}_{t+N}}_2, 
\end{equation}
where $N$ is the window size as defined in \eqref{eq:window-size}. In some applications, domain-specific knowledge might suggest that only TD (resp. FD) information is relevant for CPD. This expert knowledge can be incorporated in the dissimilarity measure by setting $\alpha=1$ and $\beta=0$ (resp. $\alpha=0$ and $\beta=1$) in \eqref{eq:combine_shared}. We denote the obtained dissimilarity measure by $\mathcal{D}_t^{\text{TD}}$ (resp. $\mathcal{D}_t^{\text{FD}}$). Using $\mathcal{D}_t^{\text{TD}}$ and $\mathcal{D}_t^{\text{FD}}$, we can also set $\alpha$ and $\beta$ automatically in such a way that the TD and FD time-invariant features contribute in a comparable fashion to $\mathcal{D}_t$. We let
\begin{align}\label{eq:alpha-beta}
\alpha = Q(\{\mathcal{D}_t^{\text{FD}}\}_t, 0.95) \quad \text{and} \quad 
\beta = Q(\{\mathcal{D}_t^{\text{TD}}\}_t, 0.95), 
\end{align}
where $Q$ is the quantile function, i.e. for a set of real numbers $A$ and $0< p \leq 1$ it holds that $Q(A,p)$ is the smallest number such that $p\cdot 100\%$ of the elements of the set $A$ are smaller than $Q(A,p)$. We use the 95-percentile as a measure of the heights of the peaks in the dissimilarity scores, where outliers are ignored. By setting $\alpha$ and $\beta$ in \eqref{eq:combine_shared} according to \eqref{eq:alpha-beta}, the peaks in $\{\mathcal{D}_t^{\text{FD}}\}_t$ and $\{\mathcal{D}_t^{\text{TD}}\}_t$ contribute approximately equally to $\{\mathcal{D}_t\}_t$. As all learned features lie in the interval $[-1,1]$, the robustness of using a quantile-based fusion approach is guaranteed.  

\subsubsection{Peak detection}
If the time-invariant features are indeed similar across successive windows within a WSS segment, the dissimilarity measure $\mathcal{D}_t$, as defined in \eqref{eq:dissimilarity-measure}, will peak at or near a change point. Determining reasonable detection alarms from these peaks is an often neglected task in current literature. In some cases, the problem is avoided by focusing on time series containing only one change point \cite{M-stat}. In other cases all local maxima of the dissimilarity measure are considered to be detection alarms \cite{Lee2018TimeLearning}, leading to unreasonably many false positives. Liu et al. \cite{Liu2013Change-pointEstimation} propose to reduce the number of false positives by deleting detections that are too close to the previous detection. As their method might also delete correct detections, it is clearly not optimal. Recently, the use of a matched filter was investigated as a way to improve detection and localization of change points \cite{cheng2020optimal, Cheng2020OnDetection}. It is however difficult to automatically select a representative peak to base the matched filter on \cite{cheng2020optimal}, nor is it possible to unambiguously derive an asymptotically matched filter \cite{Cheng2020OnDetection} for our dissimilarity measure. We therefore propose to reuse the impulse response $\mathbf{v}$ from the moving average filtering \eqref{eq:smooth-features} as it is has a comparable effect to that of a matched filter, as a consequence of its width and shape. This then leads to
\begin{equation}\label{eq:matched_filter}
    \Tilde{\mathcal{D}}_t = \sum_{k=-N+1}^{N-1} \mathbf{v}[N-k]\cdot \mathcal{D}_{t+k}.
\end{equation}
The detection alarms then correspond to all local maxima of the series $( \Tilde{\mathcal{D}}_N,  \Tilde{\mathcal{D}}_{N+1}, \ldots, \Tilde{\mathcal{D}}_{T-N})$ \cite{cheng2020optimal, Cheng2020OnDetection}. 

Aiming to further improve detection accuracy, we propose to use a different, parameter-free approach for peak detection. In topography, the \textit{prominence} of a peak is the minimum height that one needs to descend in order to be able to ascend to a higher peak \cite{llobera_2001}. The idea is that even though every peak in the dissimilarity measure might consist of multiple local maxima that all have a large height, only one of these maxima will have a large prominence. This measure has previously been successfully applied in the analysis of population data \cite{nelson_mckeon_2019}, super-resolution microscopy data \cite{griffie_boelen_burn_cope_owen_2015} and neural signals \cite{choi_ahn_park_lee_kim_cho_senok_koo_goo_2017}. Given that $\mathcal{D}_{t}$ is a local maximum, we first define the two closest time stamps left and right of $t$ for which the dissimilarity measure is larger than $\mathcal{D}_{t}$, and denote them by $t_L$ and $t_R$ respectively, i.e.,
\begin{align}
    t_L &= \max\left\{\sup\{t^* \:|\: \mathcal{D}_{t^*}>\mathcal{D}_{t} \text{ and } t^*<t\},N\right\},\\
    t_R &= \min\{\inf\{t^* \:|\: \mathcal{D}_{t^*}>\mathcal{D}_{t} \text{ and } t^*>t\},T-N\},
\end{align}
where the $\max$ and $\min$ operators ensure that $t_L$ and $t_R$ stay at a distance $N$ from the boundaries of the time series. We then define the prominence $\mathcal{P}(\mathcal{D}_{t})$ of local maximum $\mathcal{D}_{t}$ by 
\begin{equation}\label{eq:prominence}
    \mathcal{P}(\mathcal{D}_{t}) = \mathcal{D}_{t} - \max\left\{\min_{t_L<t^*<t}\mathcal{D}_{t^*},\min_{t<t^*<t_R}\mathcal{D}_{t^*} \right\}.
\end{equation}
If $\mathcal{D}_{t}$ is not a local maximum we set $\mathcal{P}(\mathcal{D}_{t})=0$ by definition. We propose to combine the matched filter \eqref{eq:matched_filter} and the prominence measure \eqref{eq:prominence}, i.e. by calculating the prominences for $\{\Tilde{\mathcal{D}}\}_t$ instead of $\{\mathcal{D}\}_t$. A change point is then detected if the prominence $\mathcal{P}(\Tilde{\mathcal{D}}_{t})$ is above a predefined threshold $\tau$. %Although this does not affect the total number of detections, it does make a difference when a particular detection threshold $\tau>0$ is chosen (cfr. Section \ref{sec:problem}). 

\subsection{Summary: the TIRE method}\label{sec:summary}

Finally, we summarize all the steps of the proposed Time-Invariant REpresentation (TIRE) change point detection method. If only time-domain or frequency-domain information is used, we will refer to the method using the acronym TIRE-TD or TIRE-FD, respectively. 

\begin{enumerate}
    \item Construct time-domain windows $\{\mathbf{y}_t\}_t$ \eqref{eq:td-windows} and frequency-domain windows $\{\mathbf{z}_t\}_t$ \eqref{eq:fd-windows} from a time series $\mathbf{X}$. 
    \item Using these windows as training data sets, train two autoencoders by minimizing loss function \eqref{eq:training-loss3}. 
    \item Use \eqref{eq:alpha-beta} to determine $\alpha$ and $\beta$ or set one of them to zero based on domain knowledge. Construct the combined time-invariant features according to \eqref{eq:combine_shared}.  
    \item Smoothen the time-invariant features according to \eqref{eq:smooth-features}. 
    \item Calculate the dissimilarity measures for all $t$ using \eqref{eq:dissimilarity-measure}. 
    \item Apply a matched filter on the dissimilarity measures following \eqref{eq:matched_filter} and compute the prominence of all local maxima using \eqref{eq:prominence}. 
    \item If the prominence \eqref{eq:prominence} of a local maximum is higher than some user-defined detection threshold $\tau$, a change point has been detected. 
\end{enumerate}

An implementation of our TIRE methods has been made available at \url{https://github.com/deryckt/TIRE}. 

%% file: experiments.tex
\section{Experiments}\label{sec:experiments-setup}
\subsection{Evaluation measure}
In our setting, the goal of a CPD algorithm is to identify the location of change points as accurately as possible. %Other interpretations of the CPD include finding segments that minimize the MSE or NLL, or classifying each time stamp as a change point or not. 
Given a toleration distance $\delta$ we say that a ground-truth change point $a$ is correctly detected by a detection alarm $b$ 
%(i.e. a peak for which the prominence (or height) is higher than the detection threshold $\tau$) 
if the following three conditions are satisfied \cite{Lee2018TimeLearning}:
%\textbf{Condition 1)} no other ground-truth change point is closer to $b$ than $a$.  \textbf{Condition 2)} the time distance between $a$ and $b$ is smaller than the toleration distance, i.e. $|a-b|\leq\delta$. \textbf{Condition 3)} the prominence (or height) of the peak corresponding to $b$ is higher than the detection treshold $\tau$. \textbf{Condition 4)} every detection alarm can only contribute to the correct detection of at most one ground-truth change point. 

\begin{enumerate}%[leftmargin=*,label={\textit{Condition \arabic*)}}]
    \item No other ground-truth change point is closer to $b$ than $a$.
    \item The time distance between $a$ and $b$ is smaller than the toleration distance, i.e. $|a-b|\leq\delta$.
    % \item The prominence (or height) of the peak corresponding to $b$ is higher than the detection treshold $\tau$.
    \item Every detection alarm can only contribute to the correct detection of at most one ground-truth change point. 
\end{enumerate}

To evaluate the performance of our method, we will construct receiver operating characteristic (ROC) curves and use the area under this curve (AUC) as a performance metric, as is common practice. Following \cite{Kawahara2012SequentialEstimation, Liu2013Change-pointEstimation, Chang2019KernelModels, Lee2018TimeLearning}, we define the true positive rate (TPR) and false positive rate (FPR) of our detection algorithm as
\begin{equation}
    \text{TPR} = \frac{N_{\text{CR}}}{N_{\text{GT}}} \quad \text{and} \quad \text{FPR} = \frac{N_{\text{AL}}-N_{\text{CR}}}{N_{\text{AL}}},  
\end{equation}
where $N_{\text{GT}}$ denotes the number of ground-truth change points, $N_{\text{AL}}$ denotes the number of all detection alarms by the algorithm and $N_{\text{CR}}$ is the number of times a ground-truth change point is correctly detected. We obtain the ROC curve by varying the detection threshold $\tau$. Unlike in the binary classification setting, the ROC curve is not necessarily monotonously increasing, as the FPR does not need to be a monotonous function of $\tau$. Nevertheless, it still holds that $0\leq \text{AUC} \leq 1$. Moreover, note that a TPR of $1.0$ can be obtained by setting the detection threshold to zero $\tau=0$ (i.e. all time stamps are detection alarms), though the FPR will always be strictly smaller than $1.0$ for $\tau=0$ when at least one change point is present. We therefore extend the ROC curve by manually adding the point $(\text{FPR},\text{TPR})=(1.0,1.0)$. This ensures that a perfect performance corresponds to an AUC of $1$. 
%In order to achieve a fair comparison of methods, we manually add the point $(1.0,1.0)$ to the curve. The reason for this is that a detection threshold of $\tau=0$ (i.e. all time stamps are detection alarms) will correspond to a TPR of $1.0$, but the FPR will be strictly smaller than $1.0$ as all ground-truth change points will inevitably be correctly detected. 
\subsection{Data sets}\label{sec:datasets}
\input{datasets}

\subsection{Parameter settings and baseline methods}\label{sec:parametersetting}

For TIRE, we report the results for two different parameter settings. Parameter setting \textit{a} corresponds to the case without instantaneous features: in both time and frequency domain the autoencoder learns only 1 (time-invariant) feature (i.e. $h^{\text{TD}}=s^{\text{TD}}=h^{\text{FD}}=s^{\text{FD}}=1$). Furthermore we set $K=2$, $\lambda^{\text{TD}}=1$ and $\lambda^{\text{FD}}=1$. Parameter setting \textit{b} corresponds to the case with 1 instantaneous and 2 time-invariant features in the time domain (i.e. $h^{\text{TD}}=3$, $s^{\text{TD}}=2$) and furthermore we set $h^{\text{FD}}=s^{\text{FD}}=1$, $K=2$, $\lambda^{\text{TD}}=1$ and $\lambda^{\text{FD}}=1$. For TIRE-TD we set $\alpha=1$ and $\beta=0$ in \eqref{eq:combine_shared}, and vice versa for TIRE-FD. For the combined approach, we set $\alpha$ and $\beta$ following \eqref{eq:alpha-beta}. We train all networks for 200 epochs using the Adam optimizer \cite{kingma2014adam} with default settings. For both parameter settings, we choose window sizes and toleration distances based on domain knowledge and sampling frequency. We set $N=20$ and $\delta=15$ for JM, SC and GM; $N=200$ and $\delta=150$ for CC; $N=10$ and $\delta=15$ for bee dance; $N=100$ and $\delta=300$ for HASC-2011 and $N=75$ and $\delta=50$ for well log. The influence of these parameter settings will be discussed in Section \ref{sec:sensitivity}. In terms of postprocessing, we use a matched filter and calculate our proposed prominence score (cf. Section \ref{sec:method}). The advantageous effect of this postprocessing stage is analyzed in Section \ref{sec:postprocessing}. In order to obtain a fair comparison, we also apply these postprocessing steps to all undermentioned baseline methods which do not explicitly define such a procedure. 

The first baseline method we use is the \textbf{generalized likelihood ratio} (GLR) procedure \cite{brandt, appel_brandt_1983}, which has been shown to have a good performance for detecting changes in the autocorrelation function or the frequency spectrum \cite{Appel1984AAlgorithms}. A conceptually similar method is described in \cite{davis1995testing}. We use a sliding window approach, where an AR(2)-model is fit on every two neighbouring windows as well as their union. A generalized log-likelihood ratio is used as dissimilarity measure. For a fair comparison, we use the same window sizes and postprocessing steps as for TIRE. 

Second, we consider a density-ratio estimation method called \textbf{relative unconstrained least-squares importance fitting} (RuLSIF) that has been applied to CPD \cite{Liu2013Change-pointEstimation}. Like with the closely related uLSIF \cite{Kawahara2012SequentialEstimation}, the idea is to estimate and compare the density ratio of two neighbouring windows instead of the individual densities. Because the validation data sets in \cite{Liu2013Change-pointEstimation} largely overlap with ours, we adopt the same parameter choices and postprocessing steps as described in the original paper.

Next, \textbf{kernel learning CPD} (KL-CPD) \cite{Chang2019KernelModels} is a recently proposed kernel learning framework for time series CPD that optimizes a lower bound of test power via an auxiliary generative model. Features are learned using a recurrent neural network and the dissimilarity measure is based on the maximum mean discrepancy. Given the large overlap in used benchmark data sets, we use the original default parameter settings in \cite{Chang2019KernelModels} without adaptation (e.g. window size of 25). We train the networks for 200 epochs, as longer training did not result in improved results. For a fair comparison, we use the same postprocessing steps as for TIRE as none were proposed in \cite{Chang2019KernelModels}. 

Finally, we compare with the \textbf{autoencoder-based breakpoint detection procedure} (ABD) \cite{Lee2018TimeLearning}. ABD only uses time-domain information and does not include any regularization to promote time-invariant features. We set parameters using the parameter guidelines in the original paper. This leads to a window size of 96 for JM, SV and GM; 995 for CC; 26 for bee dance; 158 for HASC-2011 and 155 for well log. %In addition, we expand ABD by also considering frequency-domain information and we replace the ABD parameter guidelines by our parameter settings \textit{a} and \textit{b}. To avoid confusion, we will refer to this modified ABD method as \textbf{autoencoder-based change point detection} or AECPD-(TD/FD)-\textit{a}/\textit{b}. Note that AECPD-TD only deviates from ABD in the number of encoded features and the definition of the dissimilarity measure \cite{Lee2018TimeLearning}. %By comparing AECPD and TIRE, we are able to quantify the advantage of using shared features, as the only difference between the former and the latter is the use of the shared feature penalty term in the autoencoder loss function \eqref{eq:training-loss} and the shared feature smoothing \eqref{eq:smooth-features}. 

% use the number of shared features as in parameter settings \textit{a} and \textit{b} and train the autoencoder on both TD and FD data. Note that the only difference between ABD-(TD/FD)-\textit{a}/\textit{b} and AECPD-(TD/FD)-\textit{a}/\textit{b} is the use of smoothened shared features, which enables us to exactly quantify their influence of the performance of our proposed method. For all parameter settings, the postprocessing steps from AECPD are used. 

\section{Results}\label{sec:experiments-results}

\subsection{Performance results}

In Table \ref{tab:performance}, the performances of all versions of TIRE and the baseline methods are listed. For all data sets, we report the mean AUC and its standard error. All data sets, methods and abbreviations are described in Section \ref{sec:parametersetting}. The highest mean AUC for each data set can be found in bold. In the following, we discuss some important observations. 

The GLR procedure gives very good results on the simulated data sets, but its performance degrades on the real-life data sets. This confirms the common observation that the performance of model-based CPD procedures heavily relies on how well the actual data can be described using the chosen model. In this case, both the simulated data and the GLR procedure are based on a second-order autoregressive model, which is why GLR performs well on this data. RuLSIF and KL-CPD do not perform well on data sets in which the change points manifest themselves in the frequency domain, since they do not leverage the sequential nature of the time series data, i.e. they assume the data samples to be iid. Note that AUC values for KL-CPD differ from those in \cite{Chang2019KernelModels} as CPD is there interpreted as a binary classification problem. Next, 
%even though we were able to reproduce the results of \cite{Lee2018TimeLearning} on the data sets used in the paper that are publicly available, 
ABD generally does not give good results, which can by explained by ABD's inability to detect changes in the frequency domain and the often noisy features (cf. Figure \ref{fig:representation}). In addition, ABD's \textit{normalized} dissimilarity measure (eq. (10) in \cite{Lee2018TimeLearning}), given by,
\begin{equation}
\mathcal{D}_t^{\text{ABD}} =
    %\mathcal{D}( \Tilde{\mathbf{s}}_t,\Tilde{\mathbf{s}}_{t+N}) =
    \norm{\mathbf{h}_t- \mathbf{h}_{t+N}}_2/\sqrt{\norm{\mathbf{h}_t}_2\cdot\norm{\mathbf{h}_{t+N}}_2},     
\end{equation}
where $\mathbf{h}_t$ is the vector of learned time-domain features, is not invariant to a shift of the features (i.e. adding a constant to all features); it even diverges when the norm of one of the features vanishes, which is not reasonable. 

For all data sets and both parameter settings \textit{a} and \textit{b}, either TIRE-TD or TIRE-FD outperforms (almost) all other baseline methods or has an AUC higher than $0.90$.  
%As discussed in Section \ref{sec:insight}, TIRE-FD can detect changes in the mean through the DC-component from the DFT. 
In many real-life cases, it is a priori clear whether TD (e.g. well log) or FD (e.g. HAR data, audio, \ldots) information should be used. Moreover, our framework for combining the time-invariant features from the time and frequency domain still gives consistently good results even when in one of the two domains no change point information is present. This means that the combined TD-FD approach can always be selected as a safe choice when it is unclear in which domain the change points mainly manifest themselves. Finally, the different parameter settings seem have no significant influence on the performance of TIRE. The sensitivity of the proposed method to parameter choices will be further discussed in Section \ref{sec:sensitivity}. 

\begin{table*}
\caption{Comparison of the AUC of the proposed Time-Invariant Representation CPD methods (TIRE) with baseline methods. }
\begin{center}
\begin{tabular}{|c|c|c|c|c|c|c|c|c|}
\hline
&\textbf{Mean} & \textbf{Variance}& \textbf{Coefficient}& \textbf{Gaussian} &\textbf{Bee dance} & \textbf{HASC-2011}&\textbf{Well log} & \textbf{Average}\\
\hline

GLR \cite{brandt, appel_brandt_1983}
&$0.73 \pm 0.02$ & $0.81 \pm 0.02$ & $\textbf{1.00} \pm 0.01$ & $\textbf{0.989} \pm 0.004$ & $0.55 \pm 0.06$ & $0.6431$ & $0.2109$ & $0.71 \pm 0.01$ \\ RuLSIF \cite{Liu2013Change-pointEstimation}
&$0.708 \pm 0.008$ & $0.65 \pm 0.02$ & $0.36 \pm 0.02$ & $0.874 \pm 0.007$ & $0.47 \pm 0.06$ & $0.3162$ & $0.798$ & $0.597 \pm 0.009$ \\ KL-CPD \cite{Chang2019KernelModels}
&$0.872 \pm 0.007$ & $0.23 \pm 0.02$ & $0.11 \pm 0.01$ & $0.84 \pm 0.07$ & $0.56 \pm 0.07$ & $0.343$ & $0.4247$ & $0.48 \pm 0.01$ \\ ABD \cite{Lee2018TimeLearning}
&$0.22 \pm 0.02$ & $0.17 \pm 0.02$ & $0.08 \pm 0.02$ & $0.18 \pm 0.02$ & $0.20 \pm 0.04$ & $0.2487$ & $0.477$ & $0.224 \pm 0.008$ \\ \hline TIRE-TD-\textit{a}&$0.86 \pm 0.01$ & $0.25 \pm 0.01$ & $0.26 \pm 0.01$ & $0.958 \pm 0.009$ & $0.36 \pm 0.05$ & $0.4166$ & $0.8002$ & $0.558 \pm 0.007$ \\ TIRE-FD-\textit{a}&$0.86 \pm 0.01$ & $\textbf{0.85} \pm 0.01$ & $0.96 \pm 0.01$ & $0.83 \pm 0.04$ & $\textbf{0.70} \pm 0.10$ & $\textbf{0.6504}$ & $0.6278$ & $\textbf{0.78} \pm 0.02$ \\ TIRE-\textit{a}&$0.86 \pm 0.01$ & $\textbf{0.85}  \pm 0.01$ & $0.74 \pm 0.05$ & $0.92 \pm 0.02$ & $0.65 \pm 0.09$ & $0.6172$ & $0.7656$ & $0.77 \pm 0.02$ \\ \hline
TIRE-TD-\textit{b}& $\mathbf{0.882} \pm 0.009$ & $0.26 \pm 0.02$ & $0.26 \pm 0.02$ & $0.965 \pm 0.006$ & $0.42 \pm 0.06$ & $0.4284$ & $\textbf{0.8151}$ & $0.58 \pm 0.01$ \\ TIRE-FD-\textit{b}&$0.86 \pm 0.01$ & $0.84 \pm 0.02$ & $0.95 \pm 0.02$ & $0.74 \pm 0.03$ & $0.69 \pm 0.10$ & $0.6261$ & $0.200$ & $0.70 \pm 0.02$ \\ TIRE-\textit{b}&$0.877 \pm 0.009$ & $0.83 \pm 0.02$ & $0.76 \pm 0.05$ & $0.89 \pm 0.02$ & $0.60 \pm 0.09$ & $0.6258$ & $0.8134$ & $0.77 \pm 0.01$ \\

\hline
\end{tabular}
\label{tab:performance}
\end{center}
\end{table*}

\subsection{Insight in encoded features and reconstruction}\label{sec:insight}

To gain insight into the working of the TIRE method, we investigate how the (partially) time-invariant representation and the corresponding reconstructions look like. We do this by conducting a case study on the jumping mean and bee dance data set. 

\begin{figure}
    \centering
    \includegraphics[width=\columnwidth]{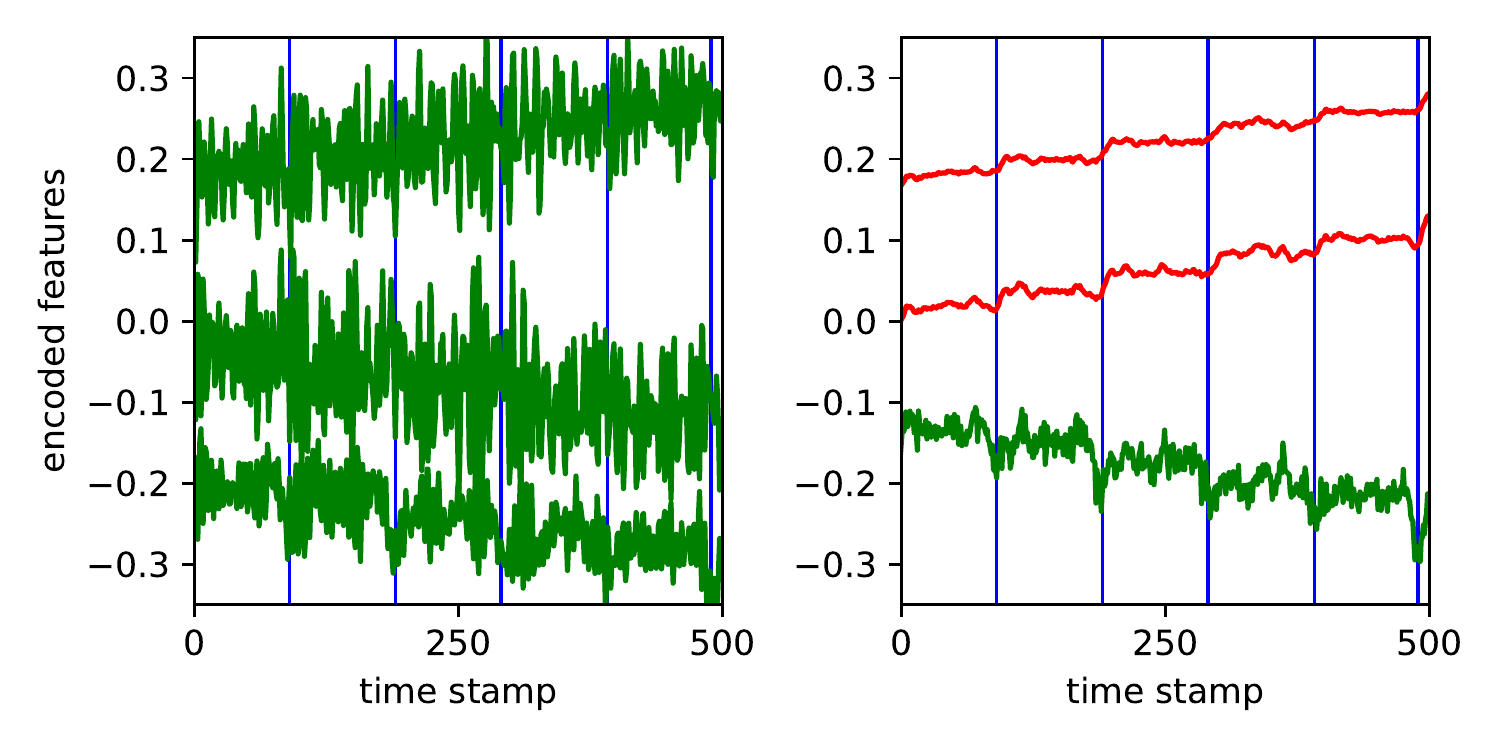}
    \caption{Example of the three-dimensional learned representation on a part of the jumping mean data set for ABD (left) and TIRE-TD (right) with two time-invariant features (in red) and one instantaneous feature (in green). The features were vertically shifted (but not rescaled) for clarity. Blue vertical lines indicate the locations of ground-truth change points. }
    \label{fig:representation}
\end{figure}

First, we demonstrate the effect of our proposed penalty in the autoencoder loss function \eqref{eq:training-loss3}. In Figure \ref{fig:representation} we show the non-smoothed encoded features (i.e. without applying \eqref{eq:smooth-features}) for a part of the jumping mean data set for both ABD and TIRE-TD. For both methods, we use three features, of which two are time-invariant in the case of TIRE. Other parameter settings are as in parameter setting \textit{b} of Section \ref{sec:parametersetting}. Whereas the features learned by ABD are very variable and noisy, the time-invariant features of TIRE-TD are approximately constant within each segment. For TIRE, the only significant variations in the features are near the ground-truth change points. These observations match exactly with the intention of our proposed loss function. 

\begin{figure}
    \centering
    \includegraphics[width=\columnwidth]{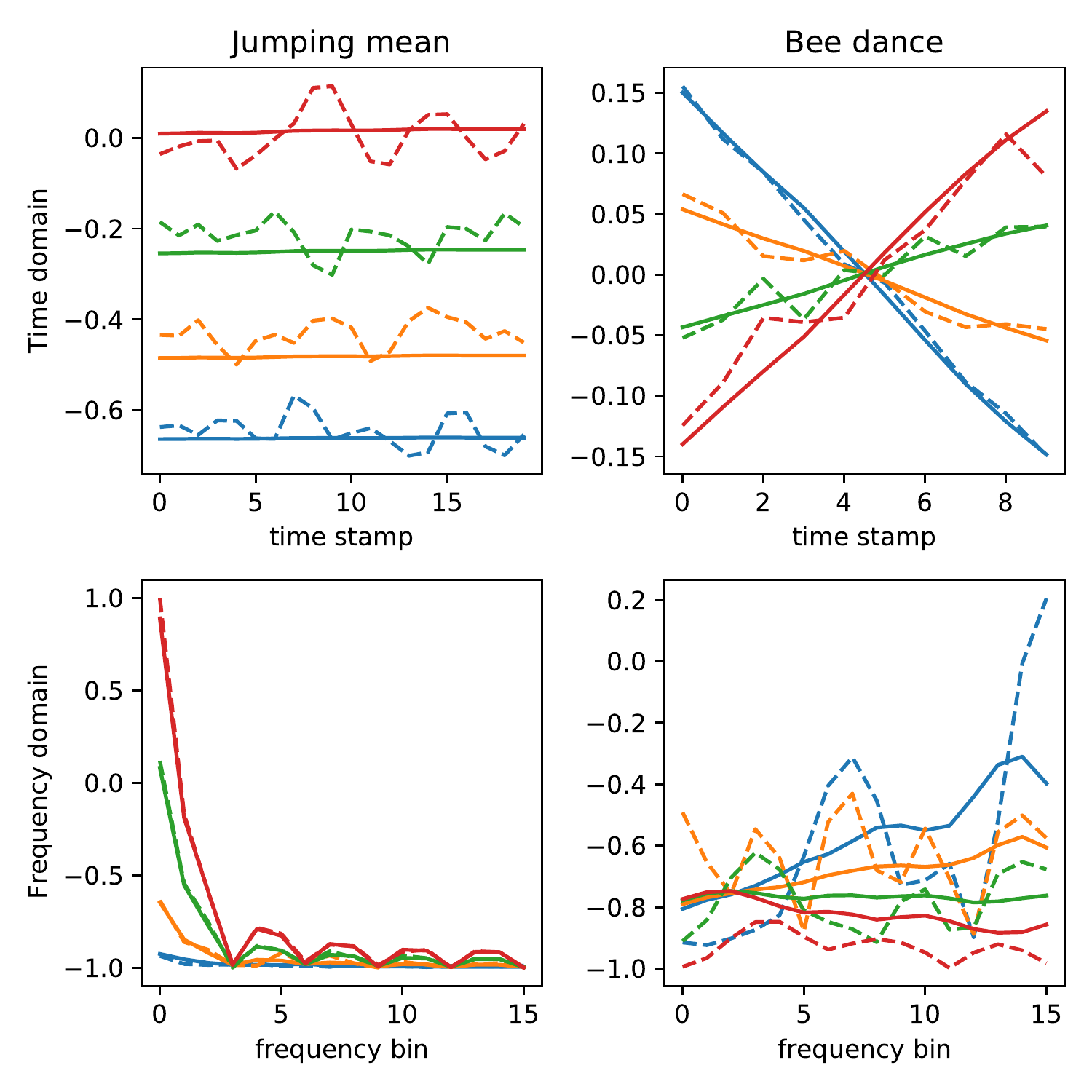}
    \caption{Examples of time-domain and frequency-domain windows (dashed lines) and their reconstructions by the autoencoder used in our proposed method (full lines). In the jumping mean data set, the change points consist of abrupt changes in mean. For bee dance, the goal is to detect abrupt changes in slope (upper right) and amplitude (lower right). }
    \label{fig:reconstruction}
\end{figure}

Second, we conduct a case study on the reconstruction of both TD and FD windows. Since the number of features we propose to use is very small, these reconstructions might be lossy and deviate from the original windows. We train TD and FD autoencoders with only one (time-invariant) feature following parameter setting \textit{a} (cf. Section \ref{sec:parametersetting}) for jumping mean and bee dance data. 
%We generate ten decoded reconstructions based on ten equidistant samples from the one-dimensional latent space of the autoencoder. 
We select four distinct windows and their reconstruction for each data set. 
The results are shown in Figure \ref{fig:reconstruction} in different colours. In case of the jumping mean data set, the autoencoder unsurprisingly reconstructs the mean of each interval, ignoring all noise. In the frequency domain, the mean manifests itself in the DC component (first frequency bin). The values of most other frequency bins seem to be encoded in the weights and biases. Next, we consider the bee dance data set. In the time domain, we use one location coordinate of the bee. As the bee moves back and forth in its waggle dance, the location coordinate resembles a triangular wave. The autoencoder can track the bees location through the variation in the slope of the location coordinate windows. The reconstruction in Figure \ref{fig:reconstruction} indeed shows approximately straight lines with varying slope. 
In the frequency domain, we only consider the angle of the head of the bee in this case study. As the bee shakes its head in some parts of the waggle dance, the goal is to pick up the presence of high-frequency oscillations. Indeed, the reconstruction only varies notably in the bins corresponding to higher frequencies. As we use only one latent variable, the decoded reconstruction does not fully capture all variations in the frequency spectrum, yet it captures the slope of the upward trend towards higher frequencies. We conclude that autoencoders can automatically identify and construct CPD-relevant features, in contrast to CPD methods based on parametric models where the relevant parameters need to be chosen in advance. 

\subsection{Importance of postprocessing}\label{sec:postprocessing}

\begin{figure}
    \centering
    \includegraphics[width=\columnwidth]{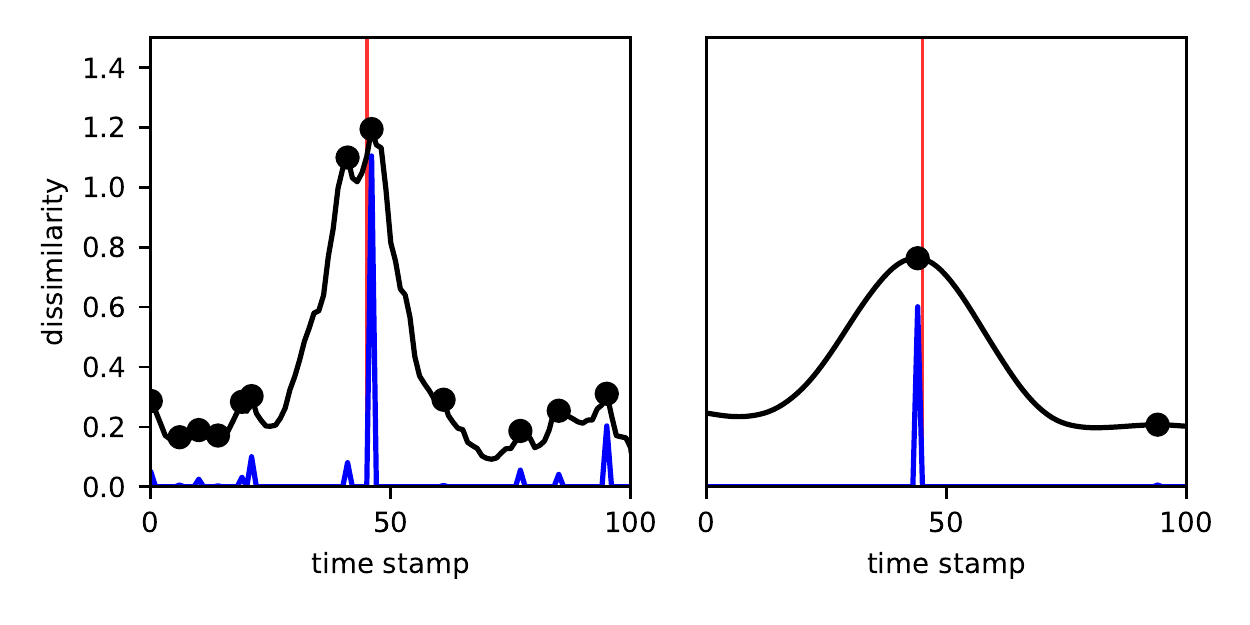}
    \caption{Example of the peak in our proposed dissimilarity measure (black line) near a ground-truth change point (red vertical line) both without (left) and with (right) the use of a matched filter. Black dots correspond to local maxima (i.e. detection alarms), our proposed prominence measure is shown in blue. The matched filter drastically reduces the number of false positive detection alarms, whereas the prominence measure makes sure that there is only one detection alarm with a large change point score. The example was generated using KL-CPD on the Gaussian mixture data set. }
    \label{fig:postprocessing}
\end{figure}

In Section \ref{sec:method}, we conjectured the importance of suitable postprocessing steps to mitigate the effect of false positive detection alarms. An example of the effect of our postprocessing steps can be found in Figure \ref{fig:postprocessing}. The use of the prominence as a change point score allows us to automatically retain only one significant detection alarm per peak, whereas a height-based dissimilarity score would lead to a false positive detection alarm even if the detection threshold is set high. Furthermore, the matched filter automatically removes most false positive detections. The use of our proposed prominence score then ensures that the remaining false positive detections have a negligible change point score. 

Next, we quantitatively compare peak height and peak prominence \eqref{eq:prominence} as change point score and investigate the effect of applying a matched filter \eqref{eq:matched_filter}. We report the average and standard deviation of the AUC on all seven data sets for the GLR procedure, RuLSIF, KL-CPD and TIRE in Table \ref{tab:postprocessing}. Both the matched filter and the use of the peak prominence result in an increase in the average AUC, with best results for when both postprocessing techniques are combined. Most notably, our proposed postprocessing approach almost leads to a doubling of the average AUC compared to naive peak detection for all methods. 

\begin{table}
\caption{Comparison of the AUC of different postprocessing techniques on dissimilarity-measure-based CPD methods. }
\begin{center}
\begin{tabular}{|c|c|c|c|c|}
\hline
&\textbf{Height} & \textbf{Height+MF}& \textbf{Prominence}& \textbf{Prom.+MF} \\
\hline
GLR&$0.42 \pm 0.05$ & $0.67 \pm 0.04$ & $0.58 \pm 0.04$ & $\textbf{0.71} \pm 0.04$ \\ RuLSIF&$0.37 \pm 0.07$ & $0.63 \pm 0.05$ & $0.60 \pm 0.07$ & $\textbf{0.64} \pm 0.05$ \\ KL-CPD&$0.28 \pm 0.10$ & $0.46 \pm 0.08$ & $0.44 \pm 0.10$ & $\textbf{0.48} \pm 0.07$ \\ TIRE&$0.40 \pm 0.08$ & $0.67 \pm 0.10$ & $0.56 \pm 0.08$ & $\textbf{0.79} \pm 0.07$ \\
\hline
\end{tabular}
\label{tab:postprocessing}
\end{center}
\end{table}

\subsection{Run time}
We compare the run times of the different methods on the jumping mean data set by reporting the mean and standard deviation of run times under 10 random seeds.
The GLR procedure takes $(6.6\pm0.4)$\si{s}, RuLSIF needs $(69.6\pm1.5)$\si{s}, KL-CPD needs $(390\pm 5)$\si{s} for 200 epochs and TIRE takes $(32.5\pm0.2)$\si{s} for 200 epochs. The run times of all methods scale linearly with the length of the time series. Unsurprisingly, the very simple GLR procedure is by far the fastest method. KL-CPD, which involves the training of a generative adversarial network and a recurrent neural network, is the slowest. Comparing the run times for 200 epochs, we see that TIRE is faster than KL-CPD. Note that the comparison between TIRE and KL-CPD is difficult, as both are iterative methods and convergence rates may differ. In the code accompanying \cite{Chang2019KernelModels}, a stop criterion for KL-CPD is provided, but this criterion was never satisfied sooner than 200 epochs on the used data sets. We conclude that TIRE has a very reasonable run time compared to other methods, albeit not the best.

\subsection{Sensitivity analysis}\label{sec:sensitivity}

We investigate to which extent the performance of the proposed method depends on the parameters chosen in Section \ref{sec:parametersetting}. Ideally, each parameter can either be set following some general guidelines, or the method should not be sensitive to the exact parameter value. 

\begin{figure}
    \centering
    \includegraphics[width=\columnwidth]{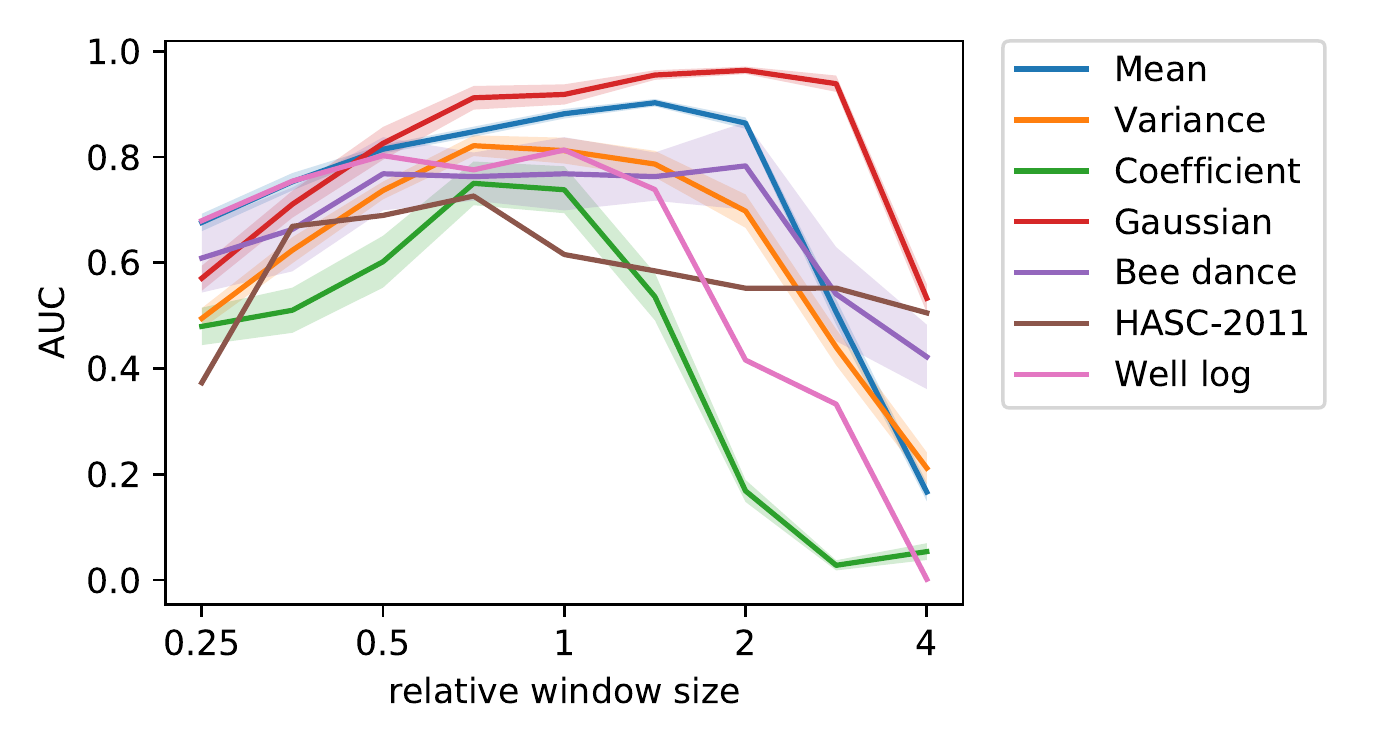}
    \caption{Influence of the window size to the performance of TIRE. We report the mean AUC and its standard error for window sizes ranging between one quarter to four times the window size chosen in Section \ref{sec:parametersetting}. }
    \label{fig:windowsize}
\end{figure}

First, we examine how the performance depends on the chosen window size. It is clear that a constant window size would in general be an unreasonable demand: when a time series is down- or upsampled, the window size should change accordingly. Some attempts to provide guidelines on how to choose a window size have been made \cite{Lee2018TimeLearning}, but these often give rise to unreasonable choices and poor performance (see ABD in Table \ref{tab:performance}). Moreover, one can even argue that a good window size is inherently dependent on the interpretation and goals of the practitioner, and can not be deduced from the data alone.  For example, this would be the case for a superposition of two CC time series (cf. Section \ref{sec:datasets}) with frequencies at two distinct scales, of which only one is of interest to the practitioner. Following amongst others \cite{Cheng2020OnDetection}, we therefore advise to set the window size based on domain knowledge (cf. Section \ref{sec:parametersetting}). To inspect the sensitivity of TIRE to these choices, we show in Figure \ref{fig:windowsize} the mean AUC and its standard error for all seven data sets for window sizes that are $0.25$, $1/2\sqrt{2}$, $0.5$, $1/\sqrt{2}$, $1$, $\sqrt{2}$, $2$, $2\sqrt{2}$ and $4$ times the domain-knowledge-based window size as defined in Section \ref{sec:parametersetting}. Furthermore we let again $K=2$, $\lambda
^{\text{TD}}=1$ and $\lambda
^{\text{FD}}=1$. 
The larger standard error for the bee dance data set in Figure \ref{fig:windowsize} is primarily caused by the large variation in difficulty between the different time series, and not by the method. For most data sets only limited variations in AUC are present in the interval $[0.5, 2]$, such that a small to moderate change in window size would not affect the positioning of the performance of the proposed TIRE method compared to the results of the considered baseline methods. For the changing coefficients (CC) data set and the well log data set, the variations in AUC are more substantial. The AUC for CC increases steadily as the window size grows since the DFT can better capture the long-range dependences in the data set, but also decreases sharply when the window size is large compared to the distance between the change points. In the well log case, the difficulty is that some change points are very close to each other. When the window size grows large, two nearby peaks in the dissimilarity measure will not be resolved anymore. In this case, the use of a matched filter is thus even disadvantageous. This also explains why the AUC decreases sharply for all data sets when an unreasonably large window size is chosen. 

Second, we investigate the influence of the latent dimension of the used autoencoder. %For the frequency-domain autoencoder, the best results were generally obtained by using one shared and zero unshared latent features. 
We let the total number of time-domain features $h^{\text{TD}}$ vary from 1 to 10 and set the number of time-invariant features to $s^{\text{TD}} = \max\{h^{\text{TD}}-1,1\}$. Furthermore we let $s^{\text{FD}}=h^{\text{FD}}=1$, $K=2$, $\lambda
^{\text{TD}}=1$ and $\lambda^{\text{FD}}=1$ (cf. parameter settings \textit{a} and \textit{b}). We use at most one instantaneous feature to avoid that the autoencoder would leak valuable CPD-relevant information into the instantaneous features (cf. Section \ref{sec:method}). We also let the number of frequency-domain features vary analogously and investigate the advantage of using time-invariant features. We do the latter by comparing to TIRE$_{\lambda=0}$, a version of TIRE with $\lambda=0$ in the loss function \eqref{eq:training-loss3} (i.e. no time-invariant features) and without the smoothing as in \eqref{eq:smooth-features}, as this is not necessarily a meaningful operation in this case. 
\begin{figure}
    \centering
    \includegraphics[width=\columnwidth]{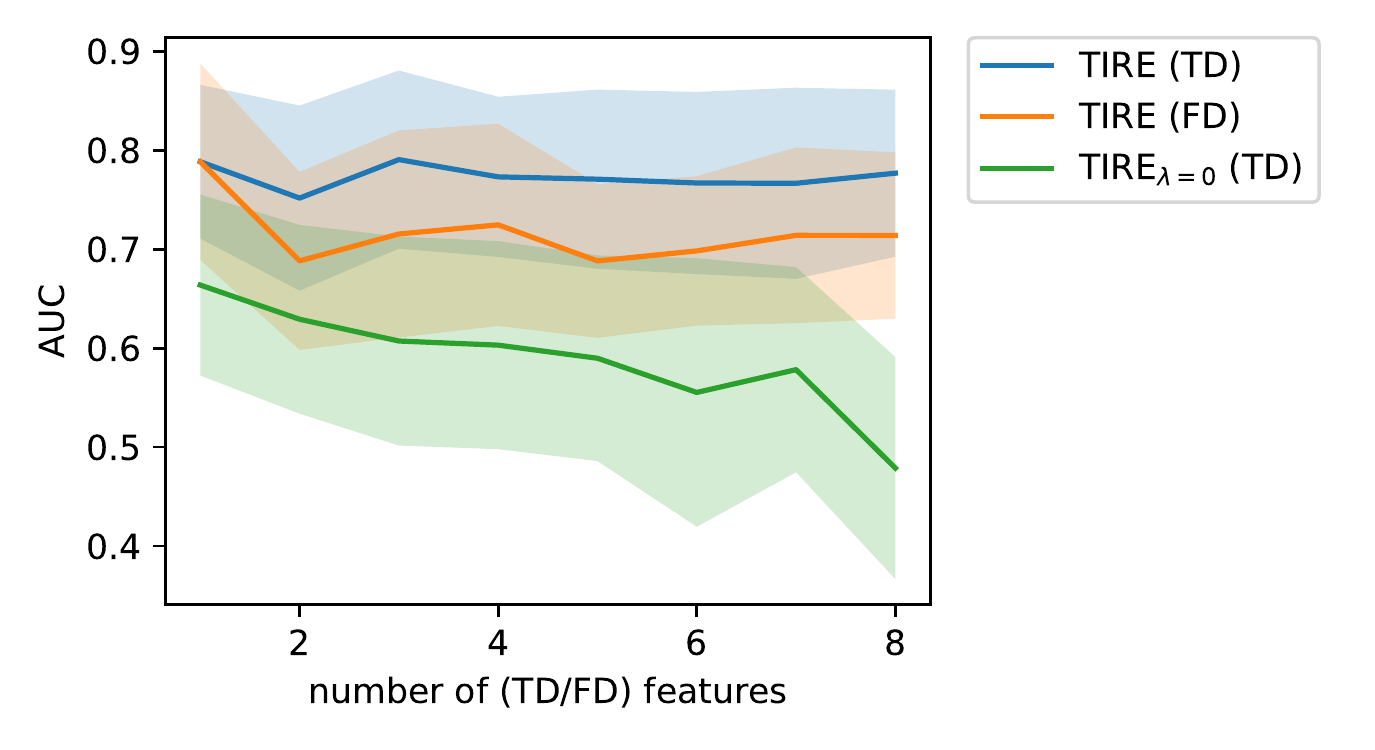}
    \caption{Sensitivity of performance of TIRE to the total number of TD features $h^{\text{TD}}$ and FD features $h^{\text{FD}}$ of the used autoencoder. The average AUC over all data sets and its standard deviation is shown. We also compare to the dependence of TIRE$_{\lambda=0}$ on the latent dimension. Whereas TIRE (with $\lambda=1$) seems on average robust to the number of TD features, the AUC for TIRE$_{\lambda=0}$ decreases. }
    \label{fig:latentdim}
\end{figure}
The average AUCs over all data sets are shown in Figure \ref{fig:latentdim}. The large standard deviation stems from the diversity of the different data sets. For TIRE, the average AUC remains very stable when the number of TD features is varied. Furthermore, the performance of TIRE seems optimal for 1 time-invariant FD feature, the average AUC when two or more FD feature are used is lower but does not further decrease with the number of FD features. 
Furthermore, we can observe that the performance of TIRE with $\lambda=1$ is clearly superior over TIRE$_{\lambda=0}$. The increase in AUC is more distinct for higher numbers of TD features. This is unsurprising, as a larger latent dimension allows an autoencoder without time-invariant features to encode the feature more freely, making the positive effect of adding the time-invariant feature term to the loss function \eqref{eq:training-loss3} all the more pronounced. 
%Finally, we quantify the advantage  of using smoothened time-invariant features by comparing AECPD and TIRE, as the only difference between the former and the latter is the use of the time-invariant feature penalty term in the autoencoder loss function \eqref{eq:training-loss} and the time-invariant feature smoothing \eqref{eq:smooth-features}. 
%Recall that the only differences between TIRE and AECPD are the use of the shared feature penalty in the loss function \eqref{eq:training-loss} and the shared feature smoothing \eqref{eq:smooth-features}. 

\begin{figure}
    \centering
    \includegraphics[width=\columnwidth]{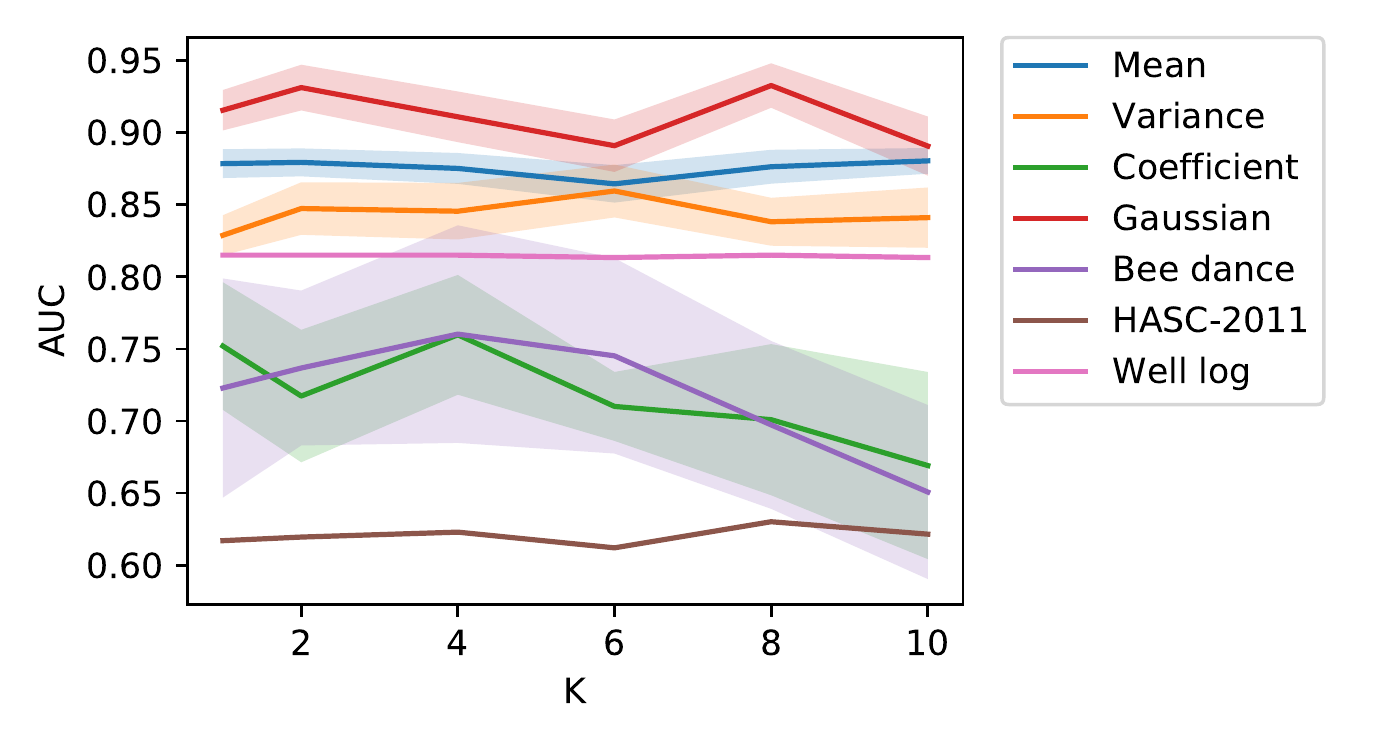}
    \caption{Sensitivity of performance of TIRE to the parameter $K$ in the TIRE training loss \eqref{eq:training-loss3}. We report for each data set the mean AUC and its standard error for $K$ between $1$ and $10$.}
    \label{fig:Kdependence}
\end{figure}

Next, we determine how sensitive TIRE is to the parameter $K$ in the training loss \eqref{eq:training-loss3}. We let $K$ vary from $1$ to $10$, with other parameters as in previous experiments, and present the result in Figure \ref{fig:Kdependence}. 
For most data sets the performance is stable with respect to changes in $K$, only for CC and bee dance a decrease in AUC is observed for large $K$. As also the runtime increases with $K$, we advise to set $K$ rather small, e.g. $K\in [1,5]$.

Finally, we investigate the sensitivity of TIRE with respect to the change magnitude at the change points (relative to the noise level). We do this by varying the standard deviation of the noise in the jumping mean data set (cf. Section \ref{sec:datasets}), leaving the change magnitudes unchanged. The jumps in the mean are of magnitude $1/16, 2/16, \ldots, 3$ and we let the standard deviation of the noise vary from $0.5$ to $3$. 
\begin{figure}
    \centering
    \includegraphics[width=\columnwidth]{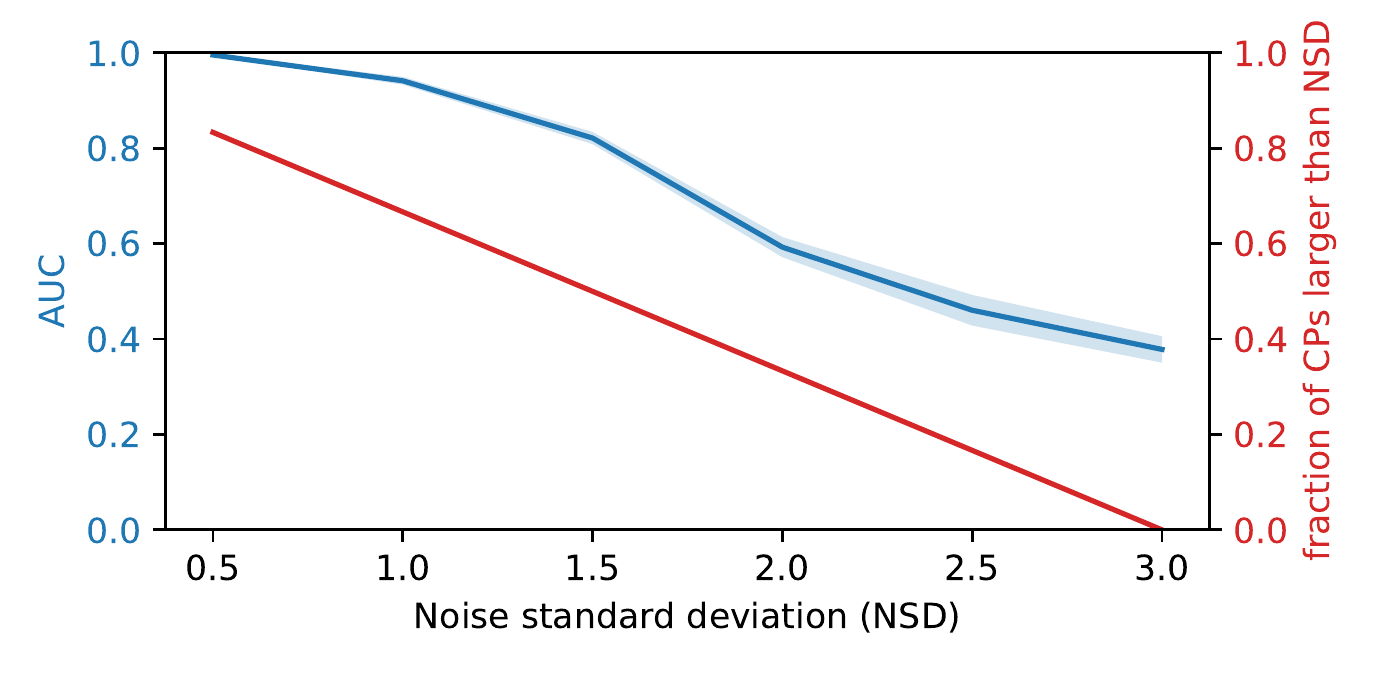}
    \caption{Sensitivity of performance of TIRE to the standard deviation of the noise in the jumping mean data set. The average AUC over ten realizations and its standard deviation is shown together with the fraction of change points for which the change magnitude is larger than the noise standard deviation.}
    \label{fig:meannoise}
\end{figure}
Figure \ref{fig:meannoise} shows a decrease of the AUC that is roughly proportionate to the fraction of change points for which the change magnitude is larger than the noise standard deviation. This is in line with expectations. 

In general, we can conclude that the performance of TIRE does not depend critically on the exact value of the window size $N$, the number of features $h$ and the parameter $K$.

%% file: datasets.tex
We demonstrate the performance of our method on four simulated and three real-life benchmark data sets, of which six are typical benchmark data sets in CPD literature. A summary of their properties can be found in Table \ref{tab:datasets}.  %We add a simulated data set based on an autoregressive model with changing coefficients as a prototypical biomedical signal \cite{rangayyan_2015}. 

\subsubsection{Simulated data}

We consider the one-dimensional autoregressive (AR) model $y(t) = a_1 y(t-1) + a_2 y(t-2) + \epsilon_t $ where $\epsilon_t\sim\mathcal{N}(\mu_t,\sigma_t^2)$ and $y(1)=y(2)=0$. We generate 50 random change points $t_n$ with $t_0=0$, $t_n=t_{n-1}+\lfloor\tau_n\rfloor$ and $\tau_n\sim\mathcal{N}(100,10)$. Following the parameter choices of  \cite{Kawahara2012SequentialEstimation, Liu2013Change-pointEstimation, Chang2019KernelModels, Takeuchi2006ASeries}, we create the following data sets, each consisting of ten randomly generated time series. 

\textbf{Jumping mean (JM)}. For this data set, let $a_1=0.6$, $a_2=-0.5$ and $\sigma_t=1.5$. We set the noise mean as
\begin{equation}
    \mu_t = \begin{cases}0 & 1\leq t\leq t_1 \\ \mu_{t_{n-1}}+n/16 & t_{n-1}+1\leq t \leq t_n. \end{cases}
\end{equation}

\textbf{Scaling variance (SV)}. For this data set, let $a_1=0.6$, $a_2=-0.5$ and $\mu_t=0$. We set the noise standard deviation as 
\begin{equation}
    \sigma_t = \begin{cases}1 & t_{n-1}+1\leq t \leq t_n \text{ and } n \text{ odd} \\ \ln(e+n/4) & t_{n-1}+1\leq t \leq t_n \text{ and } n \text{ even}.\end{cases}
\end{equation}

\textbf{Changing coefficients (CC)}. We set $a_2=0$, $\mu_t=0$ and $\sigma_t=1.5$. To take the burn-in time into account, we set $\tau_n\sim\mathcal{N}(1000,100)$. For every segment, the coefficient $a_1$ is alternatively sampled from $\mathcal{U}([0,0.5])$ and $\mathcal{U}([0.8,0.95])$, leading to clear differences in autocorrelation and frequency content between consecutive segments.
% \begin{equation}
%     a_1,t \sim \begin{cases}\mathcal{U}([0,0.5]) & t_{n-1}+1\leq t \leq t_n \text{ and } n \text{ odd} \\ \mathcal{U}([0.8,0.95]) & t_{n-1}+1\leq t \leq t_n \text{ and } n \text{ even}.\end{cases}
% \end{equation}

\textbf{Gaussian mixtures (GM)}. Here we abandon the AR model and instead simulate a piecewise iid sequence alternatively sampled between the Gaussian mixtures  $0.5\mathcal{N}(-1,0.5^2)+0.5\mathcal{N}(1,0.5^2)$ and $0.8\mathcal{N}(-1,1.0^2)+0.2\mathcal{N}(1,0.1^2)$. Change points are generated using the same mechanism as for JM and SV.
\subsubsection{Real-life data sets}
%We consider three very diverse real-life data sets. 

\textbf{Bee dance} \cite{Oh2008LearningSystems} is an often used data set to evaluate CPD algorithms \cite{Xuan2007ModelingSeries, Cheng2020OnDetection, Chang2019KernelModels, Turner2011GaussianDetection, Burg2020AnAlgorithms}. It consists of six three-dimensional time series of the bees position (location in 2D plane and angle differences) while it performs a three-stage waggle dance, which is of interest to ethnologists. 

\textbf{HASC-2011} is a subset of the HASC Challenge 2011 dataset \cite{Kawaguchi2011HASC2011corpus:Recognition}, which provides human activity data from portable three-axis accelerometers. The six activities carried out are staying still, walking, jogging, skipping, taking the stairs up or down. Following respectively \cite{cheng2020optimal} and \cite{Liu2013Change-pointEstimation}, we use the data from person 671 and convert the data to a 1D time series by taking the $l^2$-norm of the three-dimensional samples. Human activity recognition data is commonly used in CPD literature \cite{cheng2020optimal, Cheng2020OnDetection, Chang2019KernelModels, Kawaguchi2011HASC2011corpus:Recognition, Kawahara2012SequentialEstimation, Liu2013Change-pointEstimation,M-stat}.

\textbf{Well log} \cite{389767} consists of nuclear magnetic resonance measurements taken from a drill while drilling a well. Changes in the mean of the time series correspond to changes in rock stratification, outliers should be ignored \cite{Knoblauch2018Doubly-divergences}. Other results on this data set in the context of CPD evaluation include \cite{Adams2007BayesianDetection, Turner2011GaussianDetection, 389767, doi:10.1111/1467-9868.00421, Burg2020AnAlgorithms, Knoblauch2018Doubly-divergences}.

\begin{table}[htbp]
\caption{Overview of data sets. For data sets consisting of multiple time series, mean and standard deviation are reported. Q10, Q50 and Q90 denote the 10\%, 50\% and 90\% quantile, resp. }
\begin{center}
\begin{tabular}{|c|c|c|c|c|c|c|}
\hline
 & & &  & \multicolumn{3}{c|}{\textbf{CP distances}}\\
\textbf{Data set} & \textbf{Length}& \textbf{\#series}& \textbf{\#CPs} & \textbf{Q10}& \textbf{Q50}& \textbf{Q90}\\
\hline
JM, SV, GM & $4900\pm22$ & 10 & 48 & 96& 100& 104\\
CC & $49000\pm70$ & 10 & 48 & 987& 1000& 1013\\
Bee dance & $827\pm202$ & 6 & $20\pm 4$ & 28& 39& 56\\
HASC-2011 & $39397$ & $1$& $39$ & 69& 427& 2509\\
Well log & $4050$ & $1$ & $9$ & 55& 170& 390\\
\hline
\end{tabular}
\label{tab:datasets}
\end{center}
\end{table}

%% file: conclusion.tex
\section{Discussion}

In this section, we discuss some algorithmic design choices and mention potential limitations of the proposed method. 

First, the combination of time-domain and frequency-domain information is extensively studied in the field of multi-view learning \cite{zhao2017multi} and its applications. One approach is to simply concatenate separately learned TD and FD features, e.g. \cite{yuan2017multi}. Another approach is to find a joint representation, which needs to take both views into account in an effective way. This can for instance be achieved using adaptive gradient blending \cite{phan2020xsleepnet}. In the context of CPD, it is however a priori unclear how to optimize this joint representation during training. We therefore choose to train the TD and FD autoencoders separately and use a CPD-tailored data-driven weighted concatenation to fuse both views into one representation. From Table \ref{tab:performance}, it is clear that the AUC of TIRE (i.e. with TD and FD combined) is in general only slightly lower than the maximum of the AUCs of TIRE-TD and TIRE-FD, illustrating the good performance of our fusion approach. 

Second, in this paper we focused on time series with only few channels. In this setting, we showed that the latent dimension of the autoencoder has little influence on the performance. Our method deliberately targets a lossy reconstruction due to a compressed representation in order to only learn the most important time-invariant properties of the time series segment. For high-dimensional time series data, e.g. supervisory control and data acquisition (SCADA) or electroencephalography (EEG) data, the choice of latent dimension might need further investigation. Alternatively, relevant channels can be selected using an application-specific method, e.g. \cite{bertrand2020}. 

We demonstrated the performance of TIRE using the AUC, but practitioners need to choose a suitable value of $\tau$ (cf. Section \ref{sec:summary}) in order to use the method. As $\tau$ critically depends on domain knowledge and the needs of the practioner (e.g. their willingness to make a type I, resp. type II, error), we do not provide explicit guidelines. The tuning of $\tau$ can be facilitated if some prior knowledge is available, e.g. when part of the data is labelled or when an estimate of the number of change points is available. In case such information is not available and in case of doubt, we advise to underestimate $\tau$ as our proposed post-processing procedure effectively reduces the number of false positives. 

It is also worth noting that TIRE can be interpreted as a nonlinear parametric CPD method that learns the relevant parameters from the data. Whereas classical parametric methods are often able to provide an (asymptotically correct) significance level for change point probabilities \cite{davis1995testing, shao2010testing, Cheng2020OnDetection,Truong2020SelectiveMethods}, the interpretation of our change point score is rather limited. These theoretical guarantees for classical parametric methods however only hold under very specific assumptions on the data distribution, which are often not satisfied when real life data is used. 

Finally, we showed that the use of filters to both smoothen the features itself \eqref{eq:smooth-features} and the dissimilarity measure \eqref{eq:matched_filter} generally leads to a significant improvement in AUC (see Table \ref{tab:postprocessing}). Care should however be taken when the peaks in the unfiltered dissimarity measure are either skewed or very close to each other. In the first case, the peak location might shift, leading to a false negative when the toleration distance is set too small. In the second case, the two peaks might either be joined to one peak, or one of the two peaks will have a very low prominence-based change point score. Given the good performance of TIRE (Table \ref{tab:performance}), it is however clear that these are only minor concerns. 

\section{Conclusion}

We have proposed a novel distribution-free change point detection method based on autoencoders that learn a partially time-invariant representation that complies with the needs of CPD. Change points are calculated using a dissimilarity measure based on the Euclidean distance between the features learned from consecutive windows. We have mitigated the effect of false positive detections by proposing a postprocessing procedure using a matched filter and a prominence-based change point score. Furthermore, we have explicitly focused on non-iid time series by including temporally localized spectral information in the input of the autoencoder. The resulting method is very flexible, as it allows the user to indicate whether change points should be sought in the frequency domain, time domain or both. Examples of change points that can be detected are abrupt changes in the slope, mean, variance, autocorrelation function and frequency spectrum. Finally, we have showed that the performance of TIRE is consistently superior or highly competitive compared to baseline methods on benchmark data sets. A sensitivity analysis reveals that this good performance does not critically depend on the window size, nor on the latent dimension of the autoencoder. This robustness, together with the lack of distributional assumptions, make TIRE an easy-to-use change point detection method, whilst still offering a great deal of flexibility.